\documentclass{article}
\usepackage{spconf,amsmath,graphicx}
\usepackage{times}
\usepackage{helvet}
\usepackage{courier}
\usepackage{amsmath}
\usepackage{amssymb}
\usepackage{subfigure}
\usepackage{subfigure}
\usepackage{algpseudocode}
\usepackage{algorithm}

\usepackage{graphicx}
\usepackage{multirow}
\usepackage{amsfonts}
\usepackage{booktabs}       
\algdef{nSxnE}{Begin}{End}{}{}

\title{Self-supervised Adversarial Training}
\name{\normalsize {Kejiang Chen$^{\star}$\ Hang Zhou$^{\star}$\ Yuefeng Chen$^{\dagger}$\ Xiaofeng Mao$^{\dagger}$\ Yuhong Li$^{\dagger}$\ Yuan He $^{\dagger}$\ Hui Xue $^{\dagger}$\ Weiming Zhang$^{\star}$\thanks{Contact mail: zhangwm@ustc.edu.cn}\ Nenghai Yu$^{\star}$}}
\address{$^{\star}$ University of Science and Technology of China \\
	$^{\dagger}$ Alibaba Group}

\begin{document}
\ninept
\maketitle
\begin{abstract}
Recent work has demonstrated that neural networks are vulnerable to adversarial examples. To escape from the predicament, many works try to harden the model in various ways, in which adversarial training is an effective way which learns robust feature representation so as to resist adversarial attacks. Meanwhile, the self-supervised learning aims to learn robust and semantic embedding from data itself. With these views, we introduce self-supervised learning to against adversarial examples in this paper. Specifically, the self-supervised representation coupled with k-Nearest Neighbour is proposed for classification. To further strengthen the defense ability, self-supervised adversarial training is proposed, which maximizes the mutual information between the representations of original examples and the corresponding adversarial examples. Experimental results show that the self-supervised representation outperforms its supervised version in respect of robustness and self-supervised adversarial training can further improve the defense ability efficiently.
\end{abstract}

\section{Introduction}
Deep Learning has made a significant progress in computer vision, natural language processing and etc. Various kinds of techniques based on deep learning have been applied in practical engineering, such as autonomous vehicles \cite{chen2015deepdriving}, disease diagnosis \cite{li2014deep}. These empowered applications are life crucial, raising great concerns in the field of safety and security. However, recently, many studies have shown that the classifiers using neural network are not robust when encountering attacks, especially adversarial examples.

Szegedy \emph{et al.} proposed the concept of adversarial example for the first time \cite{szegedy2013intriguing}, which means that a subtle perturbation is added to the input of the neural network to produce a wrong output with high confidence. After that, plenty of methods for generating adversarial examples have been developed, including gradient-based \cite{goodfellow2014explaining,madry2017towards,dong2018boosting,kreuk2018fooling}, optimization-based \cite{szegedy2013intriguing,moosavi2016deepfool,carlini2017towards} and etc. These methods show the fragility of deep learning models. 

On the opposite side, many defenses against adversarial examples have been proposed along two directions: model hardening  \cite{goodfellow2014explaining,papernot2016distillation,kannan2018adversarial,taghanaki2019kernelized,xie2019feature}
,input preprocessing \cite{xu2017feature,song2017pixeldefend,buckman2018thermometer,jia2019comdefend,jin2019ape}. 
As for model hardening, adversarial training has been proven to be an effective defense method. One convinced reason is that adversarial training forces the neural network to learn the robust feature \cite{ilyas2019adversarial}, which is rarely affected by adversarial examples. Inspired by this view, we are eager to find neural networks that naturally learn the robust feature of images. Fortunately, self-supervised learning pursues the similar destination and has been developed quickly in recent years. Self-supervised learning aims to learn robust and semantic embedding from data itself and formulates predictive tasks to train a model, which can be seen as learning the robust representation. 

\begin{figure}
	\centering
	\includegraphics[width=3in]{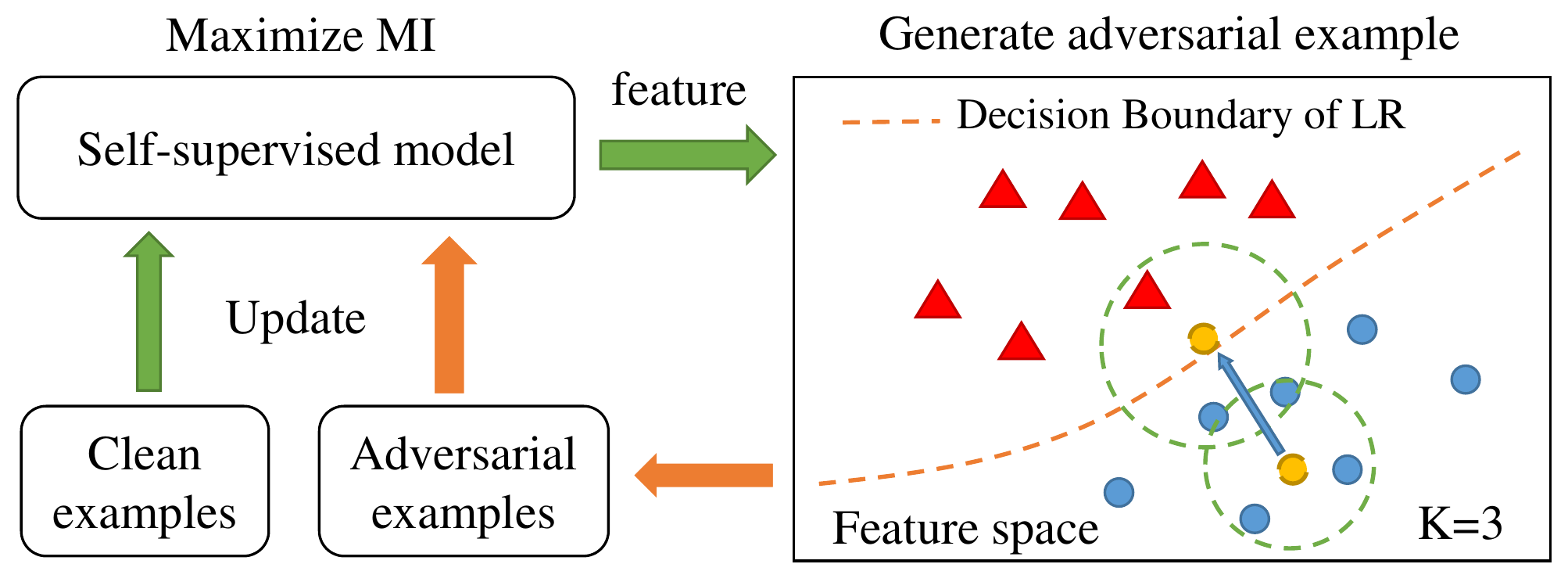}
	\caption{The diagram of self-supervised adversarial training.}
	\label{flow}
\end{figure}

Generally, given the self-supervised feature, the classification can be done with linear regression (LR) or k-Nearest Neighbors (kNN). In this paper, we choose self-supervised feature coupled with kNN as the final classifier. The reason can be intuitively observed from the right part of Fig. \ref{flow} that even the modified sample has crossed the decision boundary of LR, but it is still correctly classified by kNN, meaning that kNN owns stronger robustness than LR. 

To further enhance the robustness, self-supervised adversarial training (SAT) is proposed. The object of SAT is to maximize the mutual information (MI) between the representations of clean images and their corresponding adversarial examples, so the learned feature can mitigate the effect of adversarial perturbation. The method can be divided into two parts: generating adversarial examples, maximizing the MI. The adversarial examples are generated using gradient-based method, due to its high efficiency. Subsequently, MI between the feature representations of clean and adversarial examples is maximized. In implementation, noise contrast estimator is utilized to estimate MI. Then the model is updated by minimizing the opposite value of estimated MI. 

Our experimental results demonstrate that using the state-of-the-art self-supervised feature representation coupled with kNN shows stronger robustness against adversarial examples produced by both gradient-based and optimization-based methods with respect to supervised feature representation by a clear margin on CIFAR-10 and STL-10. Besides, the robustness of self-supervised models can be largely improved with SAT efficiently. Implementation-related file will be available at https://github.com/everange-ustc/SAT.git.

\section{Related Works}
\subsection{Adversarial Examples}
Adversarial examples are designed by an adversary to make machine learning system producing erroneous outputs. Most adversarial examples on deep neural networks are generated by adding small perturbation to clean samples. For kNN classification methods, the attack operates by adding a perturbation $\delta$ to the input such that its representation, $f(x)$, moves closer to representations of ${x}_\text{g}$, a nearest group of training instances from a different class ($x^i_g$ for $i \in \{1,2,...,m\} $). Intuitively, adversarial examples can be generated by solving the optimization problem\cite{sitawarin2019defending}:
\begin{equation}
\begin{aligned} \hat{\delta}=\underset{\delta}{\arg \min } &  \sum_{i=1}^m\left\|f\left(x_{g}^{i}\right)-f(x+\delta)\right\|_{2}^{2} \\ \text { such that } &\|\delta\|_{p} \leq \epsilon \text { and } x+\delta \in[0,1]^{d} \end{aligned}
\end{equation}
The optimization can be formulated as a Lagrangian, and we can binary search the Lagrangian constant that yields the minimal perturbation. For example, the optimization can be solved with Adam optimizer. 

\subsection{Defense}
Many defenses against adversarial examples have been proposed along two directions: model hardening, input preprocessing. For model hardening, adversarial training shows satisfying performance against adversarial examples. The \textbf{standard adversarial training (AT)} in Madry's work\cite{madry2017towards} can be formulated as:
\begin{equation}
\underset{\theta}{\arg \min }\  \mathbb{E}_{(x, y) \in \hat{p}_{\text {data }}}(\max _{\delta \in S} L(\theta, x+\delta, y))
\end{equation}
where $\hat{p}_{\text {data }}$ is the underlying distribution of training  data, $L(\theta, x, y)$ is the loss function at data point $x$ with the true label $y$ for the neural network with parameters $\theta$. $\delta$ is the permutation introduces by PGD\cite{madry2017towards}. The accuracy drops fast using AT, there is an alternate version\cite{wong2017provable}, \textbf{Mix-minibatch adversarial training (MAT)}:
\begin{equation}
\begin{aligned} \underset{\theta}{\arg \min }\ [\mathbb{E}_{(x, y) \in \hat{p}_{\text {data }}}(\max _{\delta \in S} L(\theta, x+\delta, y))+& \\\mathbb{E}_{(x, y) \in \hat{p}_{\text {data }}}(L(\theta, x, y))] 
\end{aligned}
\end{equation}
which helps to pursue the trade-off between accuracy on the clean examples and robustness on the adversarial examples.
\textbf{Adversarial logit pairing (ALP)}\cite{kannan2018adversarial} matches the logits from a clean example $x$ and its corresponding adversarial example $\tilde{x}$ during training, which exhibits better performance:
\begin{equation}
J(\mathbb{B}, \theta)+\lambda \frac{1}{m} \sum_{i=1}^{m} L\left(f\left(x ; \theta\right), f\left(\tilde{x} ; \theta\right)\right)
\end{equation}
where $\mathbb{B}$ is a minibatch including clean examples $x$ and the corresponding adversarial examples $\tilde{x}$. $f(x;\theta)$ is function mapping from inputs to logits of the model and $J(\mathbb{B}, \theta)$ is the cost function used for adversarial training. One potential reason of adversarial training is that it forces the neural network to learn robust feature, which can mitigate the affect of adversarial examples\cite{ilyas2019adversarial}. 

\subsection{Self-supervised Learning}
Self-supervised learning exploits internal structures of data and formulates predictive tasks to train a model, which can be seen as learning the robust feature. Here are some representative works in this aspect: Contrastive Predictive Coding (CPC) \cite{oord2018representation} uses a probabilistic contrastive loss which induces the latent space to capture information that is maximally useful to predict future samples. Deep Infomax (DIM) \cite{hjelm2018learning} maximizes mutual information between global features and local features. Augmented Multiscale DIM (AMDIM) \cite{bachman2019amdim} maximizes mutual information between features extracted from multiples views of a shared context.

Actually, the self-supervised representation has been used for defense in previous works. \cite{sitawarin2019defending} utilized the feature representation coupled with kNN for classification. Both supervised and self-supervised features are adopted. However, as mentioned in \cite{sitawarin2019defending}, their method does not perform well on datasets bigger than MNIST.  \cite{hendrycks2019using} combined the self-supervised loss into the loss of traditional adversarial training, but this training process is still time-consuming.

\section{Method Description}
As illustrated before, forcing the neural network to learn the robust feature of the instance can help improve the robustness of the model. Meanwhile, the self-supervised learning focuses on the robust feature, for example. they can predict the missing part of images using itself. Inspired by these point-views, we propose using self-supervised representation cooperated by k-Nearest Neighbour for defending against adversarial examples. Besides, we can maximize the mutual information representation between clean and adversarial examples by adjusting the existing model, so that the model can further mitigate adversarial perturbation.
%
%

\subsection{Self-supervised Representation for Defense}
The self-supervised representation is coupled with kNN for classification. After self-supervised training, the neural network is frozen and adopted as a feature extractor. All instances in the training set are fed into the network to obtain their representations on a specified layer, and then these representations serve as the feature library. Given an image, extract its feature representation, search the k-nearest representations from the feature library, and then predict the label of the image.

\subsection{Self-supervised Adversarial Training}


To further improve the robustness of self-supervised representations cooperated with kNN, we propose a method called self-supervised adversarial training (SAT), which maximizes the mutual information between the representations of clean images and the corresponding adversarial examples. As shown in Fig. \ref{flow}, given the pretrained self-supervised model, the framework of SAT is divided into two parts: generating adversarial examples and maximizing the mutual information.

\subsubsection{Generating Adversarial Examples}
Due to the introduced attack method in Section 2.1 is time-consuming, we modified the generating method inspired by PGD\cite{madry2017towards}. In detail, the gradient of the image is obtained firstly:
\begin{equation}
{g} = \nabla _{ {x}^{t-1}_{\text{adv}}}\sum_{i=1}^m{\lVert f\left(  {x}^i_\text{g} \right) -f\left(  {x}^{t-1}_{\text{adv}} \right) \rVert _{2}^{2}}
\end{equation}
where $\nabla$ is the gradient operator, and $m=300$ is the default setting. Then update the image:
\begin{equation}
{x}_{\text{adv}}^{t}= {x}_{\text{adv}}^{t-1}-\epsilon _s\cdot \text{sign}( {g})
\end{equation}
where $\epsilon_s$ is the update step size. To restrict the generated adversarial examples within the $\epsilon$-ball of $ {x}_\text{adv}$, we can clip $ {x}_\text{adv}$ after each update.
For better distinction, we address the former method in Section 2.1 as optimization-based method and this as gradient-based method.

\subsubsection{Maximizing Mutual Information}
After obtaining the adversarial examples, we are going to maximize the MI on the feature representation space. 
Formally, the MI between $X$ and $X_\text{adv}$, with joint density $p(x,x_\emph{adv})$ and marginal densities $p(x)$ and $p(x_\emph{adv})$, is defined as the Kullback–Leibler (KL) divergence between the joint and the product of the marginals:
\begin{equation}
\begin{array}{ll}
I(X ; X_\text{adv}) &=D_{\mathrm{KL}}(p(x, x_\emph{adv}) \| p(x) p(x_\emph{adv}))\\
&=\mathbb{E}_{p(x, x_\emph{adv})}\left[\log \frac{p(x, x_\emph{adv})}{p(x) p(x_\emph{adv})}\right]
\end{array}
\end{equation}
As for the feature representation, the MI can be defined as:
\begin{equation}
I\left(z_{i} ; \hat{z_i}\right)=\underset{z_{i}, \hat{z_{i}}}{\mathbb{E}}   \left[  \log \frac{p\left(z_{i}, \hat{z_{i}}\right)}{p\left(z_{i}\right) p\left(\hat{z_{i}}\right)}\right]
\end{equation}
where $z_{i},\hat{z_i}$ are the feature representations of clean images and the corresponding adversarial version, respectively. It is hard to obtain the explicit distribution of representations, meaning that the MI cannot be calculated. Instead, several methods have been proposed to estimate MI, and here noise contrast estimator (NCE) is adopted, whose estimated MI has been proved to be a low bound of MI\cite{tschannen2019mutual}, defined by:
\begin{equation}
\begin{array}{ll}
I(Z ; \hat{Z}) &\geq \mathbb{E}\left[\frac{1}{N} \sum_{i=1}^{N} \log \frac{\Phi \left( \left\{ z_i,\hat{z}_i \right\} \right)}{\frac{1}{N} \sum_{j=1}^{N} {\Phi}\left( \left\{ z_i,\hat{z}_j \right\} \right) }\right] \\ & \triangleq I_{\mathrm{NCE}}(Z ; \hat{Z})
\end{array}
\label{nce}
\end{equation}
where  $\hat{z}_j$ is the representation of other adversarial example different from $\hat{z}_i$. Here, we refer to representations from joint distribution as positives, i.e. $\emph{pos} \sim p\left(z_i, \hat{z_i}\right)$, $\emph{pos}=\{z_i,\hat{z_i}\}$, and representations from the product of marginal distributions as negatives, i.e. $\emph{neg} \sim p\left(z_i\right) p\left(\hat{z_j}\right)$, $\emph{neg}=\{z_i,\hat{z_j}\}$. $N$ in in Equation (\ref{nce}) is the number of negative pairs, and $\Phi(\cdot)$  is the score function that is higher for positive pairs but lower for negative pairs. $\Phi(\cdot)$ can be any continuous and differentiable parametric functions, such as cosine similarity function. Here, the matching score function is defined as a simple dot product:
\begin{equation}
\Phi \left( z_i,\hat{z}_i \right) \triangleq \phi _1\left( z_i \right) ^{\top}\phi _2\left( \hat{z}_i \right) 
\end{equation}
where $\phi _1(\cdot)$ and $\phi _2(\cdot)$ are small neural networks, for they can approximate any superb score functions.
In implementation, the estimated MI is maximized by minimizing its opposite value, named the contrast loss:
\begin{equation}
\mathcal{L}_{\text{contrast }}=-\underset{\left\{ z_i,\hat{z}_i \right\}}{\mathbb{E}}\left[ \frac{1}{N} \sum_{j=1}^{N} \log \frac{\Phi \left( \left\{ z_i,\hat{z}_i \right\} \right)}{\sum_{j=1}^N{\Phi}\left( \left\{ z_i,\hat{z}_j \right\} \right)} \right] 
\label{contrast}
\end{equation}
The self-supervised neural network can be fine-tuned using back-propagation through minimizing $\mathcal{L}_{\text{contrast }}$. The process will be kept iterating until the performance meeting the requirement.
To point out, the whole process does not require the true label of data, similar to the self-supervise learning.
The pseudo-code of the framework is given in \textbf{Algorithm 1}.

\begin{algorithm}[t]
	\caption{
		Self-supervised Adversarial Training (SAT)
	}
	\label{alg:adv_train}
	\begin{algorithmic}[1]
		\Require Training samples $X$, perturbation bound $\epsilon$, step size $\epsilon_s$, maximization iterations per minimization step $K$, and minimization learning rate $\tau$.
		\State Initialize $\theta$ with a pretrained self-supervised model $f$.
		\For{epoch $= 1 \ldots N_{ep}$}
		\For{minibatch $B\subset X$}
		\State Build $x_{adv}$ for $x \in B$:
		\Begin
		\State Assign a random perturbation
		\State \qquad $r \gets U(-\epsilon,\epsilon)$
		\State \qquad $x_{adv} \gets x + r$
		\For{ $k = 1 \ldots K$}
		\State $L \gets \sum_{i=1}^m\left\|f\left(x_{g}^{i}\right)-f(x_{adv})\right\|_{2}^{2}$
		\State $g_{adv} \gets \nabla _{ {x}_{\emph{adv}}} L$
		\State $x_{adv} \gets x_{adv} - \epsilon_s \cdot \text{sign} (g_{adv})$
		\State $x_{adv} \gets \text{clip}(x_{adv}, x-\epsilon, x+\epsilon) $
		\EndFor
		\End \vspace{1mm}
		
		\State Calculate the representation of samples:
		\Begin
		\State    $\hat{z} \gets f(x_{adv}),z \gets f(x)$
		\End
		\State Update $\theta$ with stochastic gradient descent:
		\Begin
		\State  $g_\theta \gets  \mathbb{E}_{x \in B}  [\nabla_\theta \, \mathcal{L}_{\text{contrast }}(z,\hat{z})]$
		\State  $\theta \gets \theta - \tau g_\theta$ 
		\End
		\EndFor
		\EndFor
	\end{algorithmic}
\end{algorithm}

\section{Experiments}
\subsection{Setting} 
\quad\ \ \textbf{Dataset} CIFAR-10 and STL-10 are selected as the dataset. 
The CIFAR-10 dataset consists of 60000 $32\times32$  labeled color images in 10 classes, with 6000 images per class. There are 50000 training images and 10000 test images.
STL-10 is composed of 10 classes 5000  $96 \times 96$ labeled color images, 100000 unlabeled images for training and 8000 labeled images for testing. For speedy training, we resize the images in STL-10 to $64 \times 64$. 

\textbf{Attack Method}
The attack is implemented under white-box setting: the attacker has full information about the model (i.e. knows the architecture, parameters, etc.). Both gradient-based and optimization-based attack methods are utilized to evaluate the robustness of models. All the adversarial examples are generated on the 1000 correctly predicted images on the testing set.
For the gradient-based attack methods, there are two kinds of setting: $\epsilon=0.03$, $\epsilon_s=0.005$, and 10 iterations; $\epsilon=0.06$, $\epsilon_s=0.005$, and 20 iterations, which is denoted as \textbf{small} perturbation and \textbf{large} perturbation, respectively. 

\textbf{Evaluation Metric}
For kNN classification, $k$ is 75 and faiss\cite{JDH17} is adopted for speed consideration. The penultimate layer of neural network is adopted as the representation. The defense successful rate (\textbf{DSR}), defined as the correct prediction rate on the adversarial examples, is utilized to evaluate the robustness of the model against gradient-based attack. $\ell_2$ distance, the average $\ell_2$-norm of perturbation required to mislead the classifier, is used to measure the robustness against optimization-based attack. Larger $\ell_2$ distance leads to better robustness. The accuracy (\textbf{ACC}) of clean examples is also presented to show the precision of the model.

\begin{table}[]
		\caption{The defense results of self-supervised representation and supervised representation of AMDIM with kNN on CIFAR-10 and STL-10.}
	\begin{tabular}{@{}cccccc@{}}
		\toprule
		\multirow{2}{*}{Dataset} & \multirow{2}{*}{Method} & \multirow{2}{*}{ACC} & \multicolumn{2}{c}{DSR} & \multirow{2}{*}{$\ell_2$ distance} \\ \cmidrule(lr){4-5}
		&  &  & Small & Large &  \\ \midrule
		\multirow{2}{*}{CIFAR-10} & SUP & \textbf{92.02\%} & 18.7\% & 15.4\% & 0.378 \\
		& SSL & 84.64\% & \textbf{51.9\%} & \textbf{27.7\%} & \textbf{0.667} \\ \cmidrule(l){1-6} 
		\multirow{2}{*}{STL-10} & SUP & 75.41\% & 24.2\% & 16.3\% & 0.970 \\
		& SSL & \textbf{86.13\%} & \textbf{54.9\%} & \textbf{44.7\%} & \textbf{1.591}\\
		\bottomrule
	\end{tabular}
    \label{supvsssl}
\end{table}
\begin{figure}[t]
	\centering
	\subfigure[AMDIM]{\includegraphics[width=1.65in]{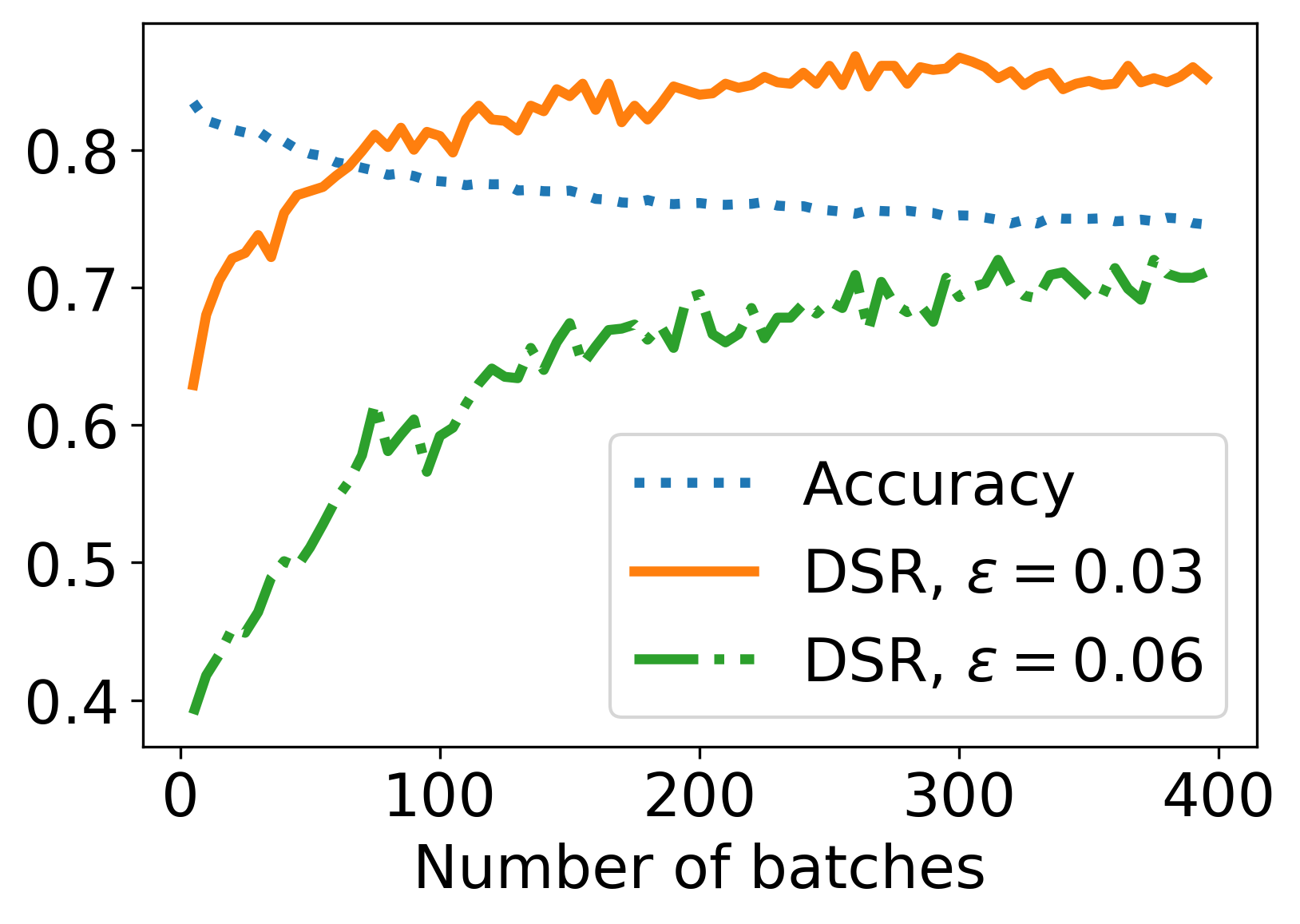}}
	\subfigure[NPID]{\includegraphics[width=1.65in]{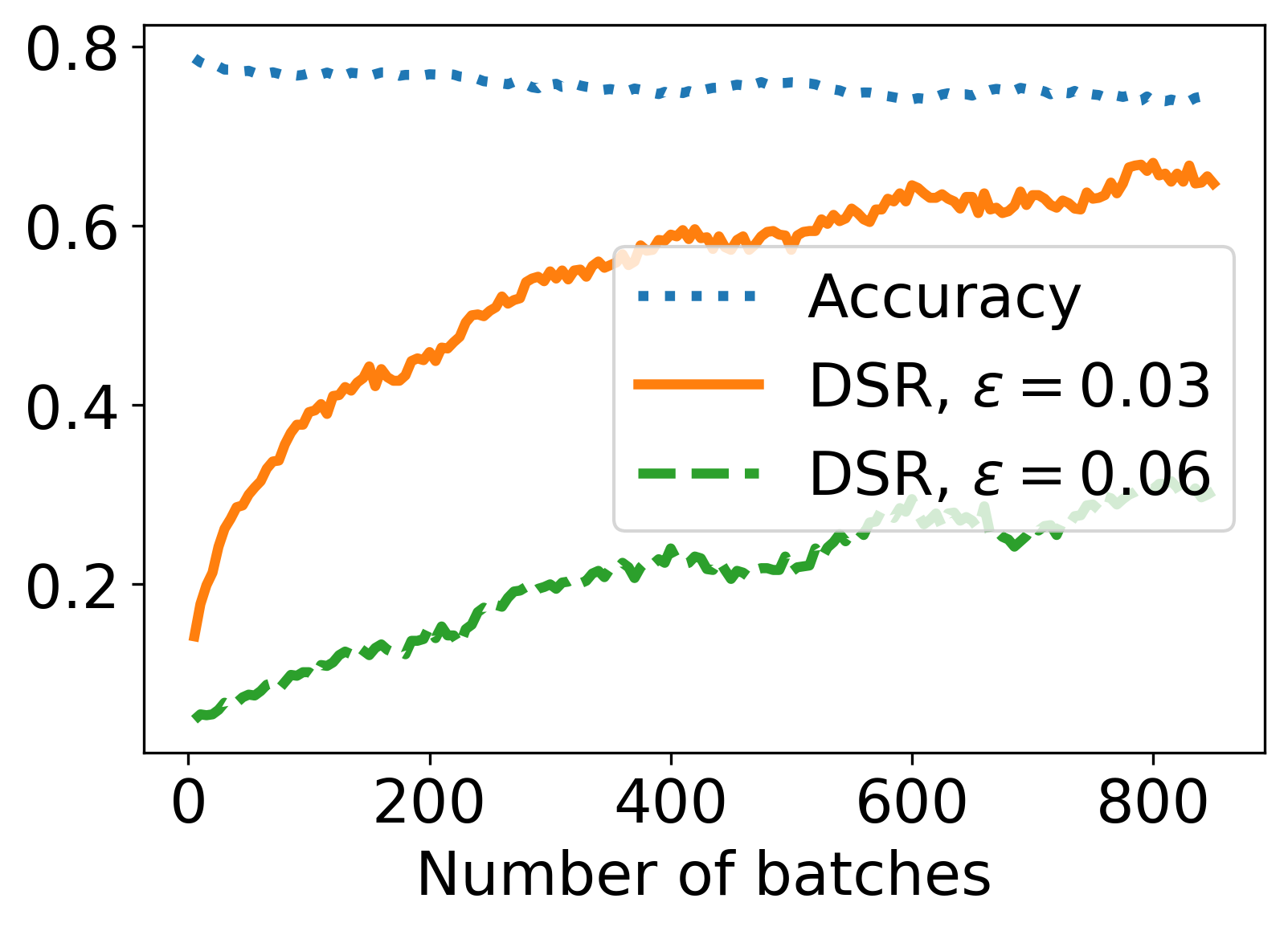}}
	\caption{The defense results of AMDIM and NPID using SAT on CIFAR-10.}
	\label{batch}
\end{figure}

\begin{figure}[t]
	\centering
	\subfigure[Small perturbation]{\includegraphics[width=1.65in]{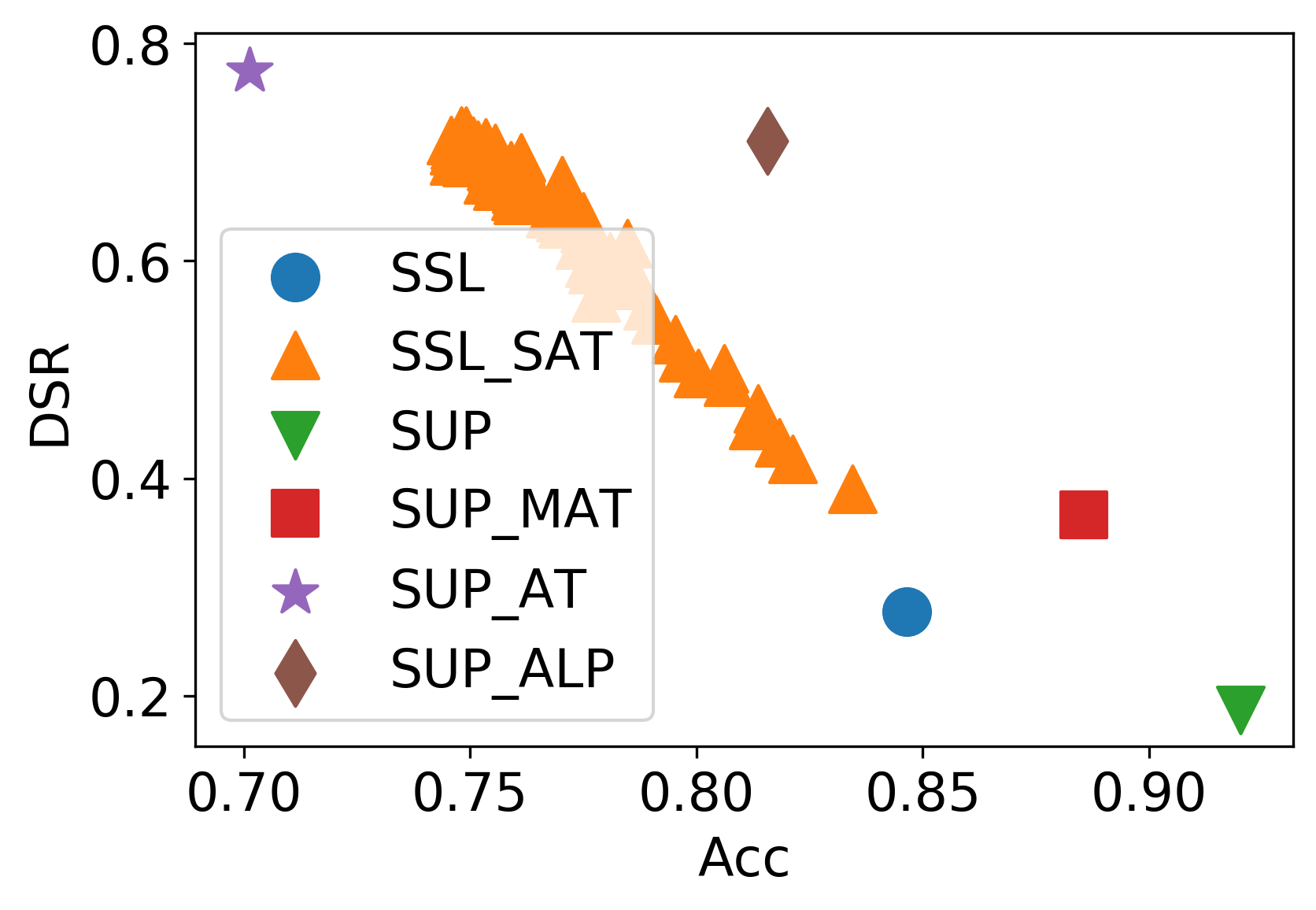}}
	\subfigure[Large perturbation]{\includegraphics[width=1.65in]{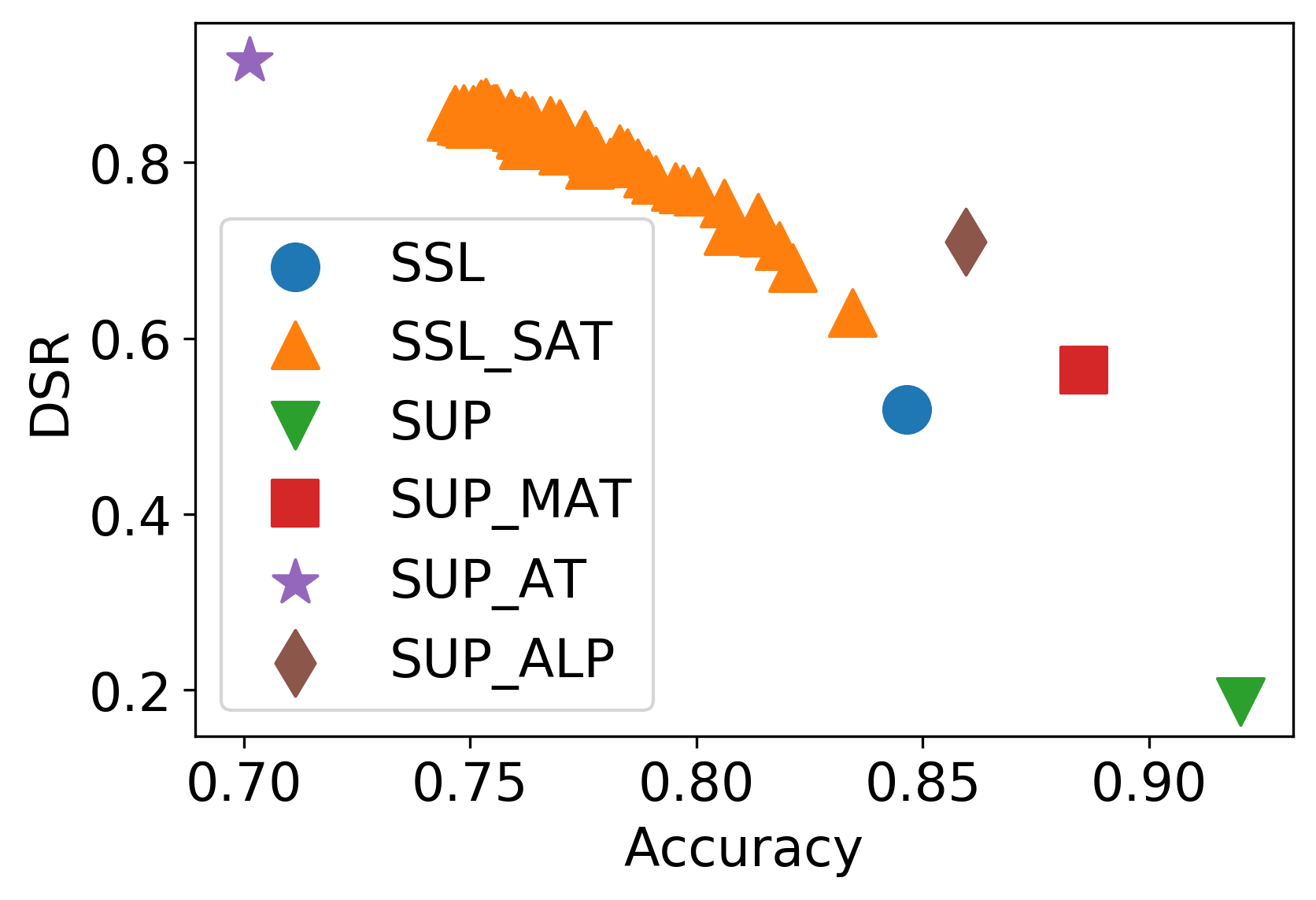}}
	\caption{The defense results of among self adversarial training and supervised adversarial training on CIFAR-10. AMDIM is selected as the seed model, and SUP and SSL mean the supervised and self-supervised version.}
	\label{cifar10_003}
\end{figure}
\begin{figure}[t]
	\centering
	\subfigure[Small perturbation]{\includegraphics[width=1.65in]{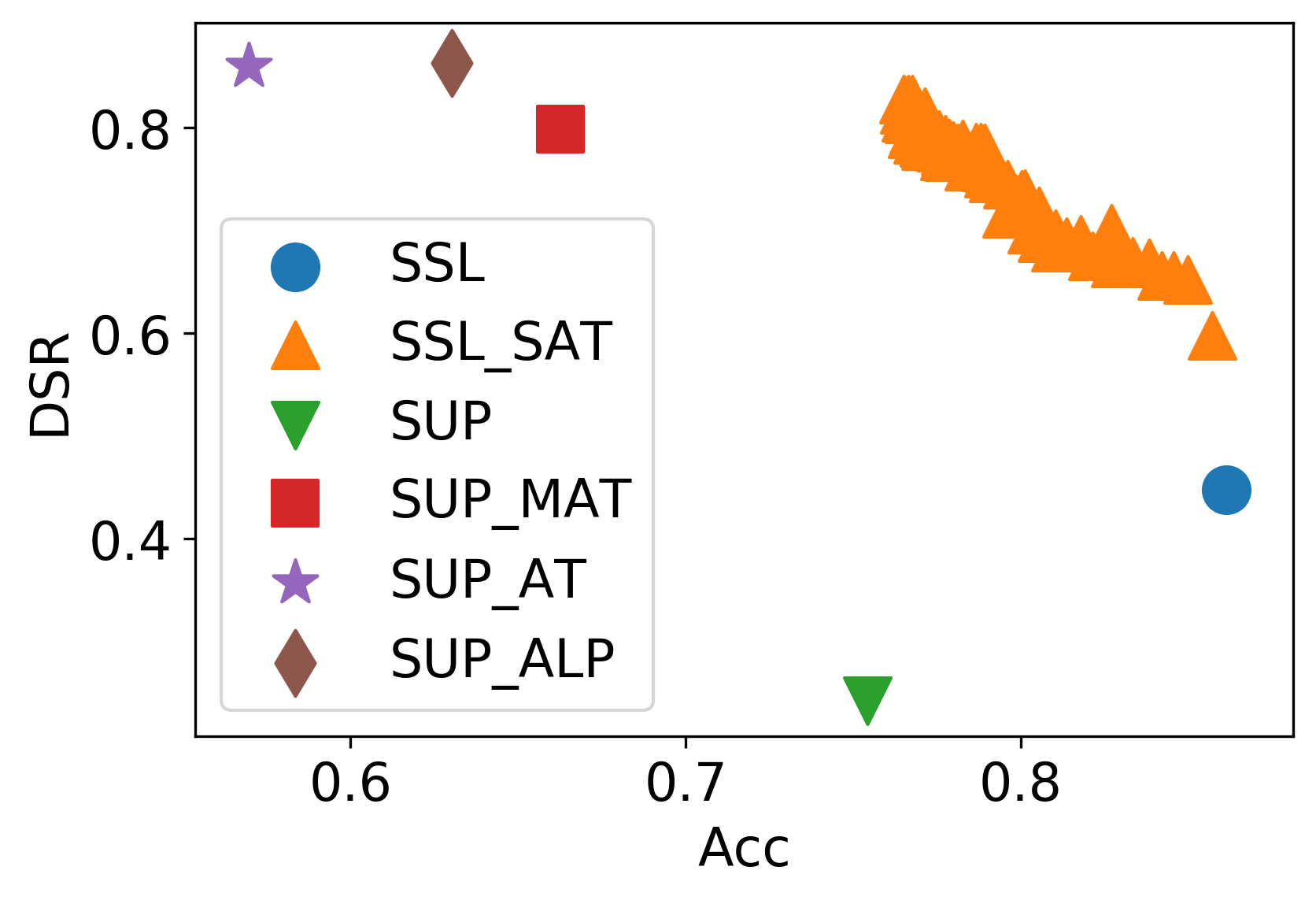}}
	\subfigure[Large perturbation]{\includegraphics[width=1.65in]{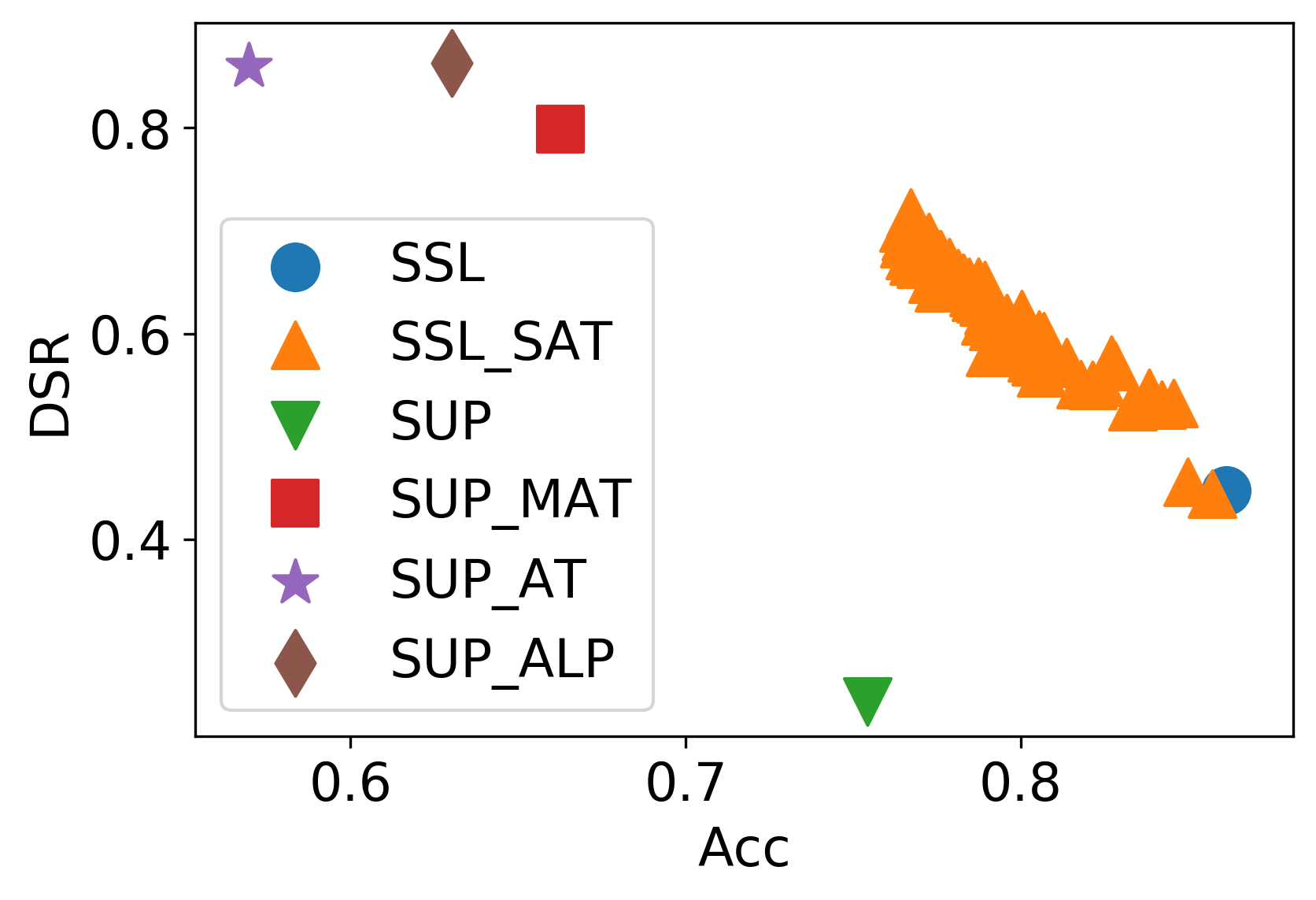}}
	\caption{The defense results of among self adversarial training and supervised adversarial training on STL-10.}
	\label{cifar10_006}
\end{figure}

\subsection{Superior of Self-supervised Representation}
In this experiment, the state-of-the-art self-supervised learning method, AMDIM \cite{bachman2019amdim}, and its supervised versions are compared to present the superior of self-supervised feature representation in the respect of robustness. These two methods are denoted by SSL and SUP, respectively. The backbone of AMDIM is an encoder based on the standard ResNet\cite{he2016deep}, with changes to make it suitable for DIM. More details about the encoder, the readers can refer to \cite{hjelm2018learning}. The parameters of the encoder for CIFAR-10 and STL-10 are set as (ndf=128, nrkhs=128, ndepth=3), (ndf=128, nrkhs=1024, ndepth=8), respectively. 
For supervised learning, Adam optimizer with learning rate 0.001 is adopted, and we trained the model for 400 epochs. For self-supervised learning, the learning rate is 0.0002 and the number of epoch is 300.

As shown in Table \ref{supvsssl}, the DSR of the self-supervised learning model (SSL) of AMDIM outperforms its supervised learning model  (SUP) with a clear margin against gradient-based attack and the required $\ell_2$-distance of successfully attacking SSL is larger than that of the SUP on two datasets. On CIFAR-10, the ACC on the clean images is lower than its supervised version, but the gain on the DSR is large, 33.2\% for small perturbation attack. On STL-10, the performance is considerable, whose ACC outperforms that of supervised version, benefiting from unlabeled images in the training phase. In conclusion, the self-supervised representation owns stronger robustness.

\subsection{Effectiveness of SAT}
To verify the effectiveness of SAT, unsupervised model NPID\cite{wu2018unsupervised} and self-supervised model AMDIM\cite{bachman2019amdim} are selected as the seed models. The adversarial examples are generated by gradient attack method with the small perturbation. Batch size 100, learning rate 0.0001 and Adam optimizer are the other setting for SAT. The results are presented in Fig. \ref{batch}. It can be seen that the DSR improves a lot after SAT with slight drop of accuracy for both AMDIM and NPID against small and large perturbation attacks, verifying the effectiveness of SAT.

We have also compared with the supervised adversarial training methods, and the adversarial examples are generated using PGD with the same setting as the gradient-based method does. AMDIM is selected as the seed model, and the results of these methods against small and large perturbation attack are shown in Fig. \ref{cifar10_003}, Fig. \ref{cifar10_006}, and the suffix represents which adversarial training method is adopted. The closer to the top right corner in figures, the better the performance. For sufficient label dataset CIFAR-10, the SAT is worse than the MAT, ALP, due to the original classification performance of self-supervised learning is worse than supervised models. Analyzing the development of self-supervised learning, we can see the gap between self-supervised and supervised model is closer. With stronger self-supervised model, the performance of SAT will become considerable. Furthermore, the time cost of SAT is much cheaper than that of MAT and ALP. For dataset with a few labels, like STL-10, the performance of SAT is significant. The trade-off between the robustness and accuracy is better achieved by SAT than supervised versions, especially for small perturbation attack, shown in Fig. \ref{cifar10_006}. Since the dataset with a little supervised information is common in many downstream tasks, the proposed method SAT has a good prospect.

\section{Conclusion}
In this paper, we utilize self-supervised representation coupled with kNN for classification, where the underlying reason is that self-supervised model learns the robust feature of data. To further strengthen the defense ability of self-supervised representation, a general framework called self-supervised adversarial training is proposed, which maximizes the mutual information between the representations of original examples and adversarial examples.  The experiments show that the self-supervised representation of AMDIM outperforms its supervised representation in the aspect of  robustness on CIFAR-10 and STL-10. Furthermore, self-supervised adversarial training has been verified that it can be efficiently applied to AMDIM and NPID, and significantly improve the robustness against adversarial examples with slight drop of accuracy.

 It is interesting to design self-supervised learning which considers adversarial attack in the training phase, so that the self-supervised representation naturally owns strong robustness, which is one direction of our future work.

\clearpage
\bibliographystyle{IEEEbib}
\bibliography{refs}

\begin{thebibliography}{10}

\bibitem{chen2015deepdriving}
Chenyi Chen, Ari Seff, Alain Kornhauser, and Jianxiong Xiao,
\newblock ``Deepdriving: Learning affordance for direct perception in
  autonomous driving,''
\newblock in {\em Proceedings of the IEEE International Conference on Computer
  Vision}, 2015, pp. 2722--2730.

\bibitem{li2014deep}
Rongjian Li, Wenlu Zhang, Heung-Il Suk, Li~Wang, Jiang Li, Dinggang Shen, and
  Shuiwang Ji,
\newblock ``Deep learning based imaging data completion for improved brain
  disease diagnosis,''
\newblock in {\em International Conference on Medical Image Computing and
  Computer-Assisted Intervention}. Springer, 2014, pp. 305--312.

\bibitem{szegedy2013intriguing}
Christian Szegedy, Wojciech Zaremba, Ilya Sutskever, Joan Bruna, Dumitru Erhan,
  Ian Goodfellow, and Rob Fergus,
\newblock ``Intriguing properties of neural networks,''
\newblock {\em arXiv preprint arXiv:1312.6199}, 2013.

\bibitem{goodfellow2014explaining}
Ian~J Goodfellow, Jonathon Shlens, and Christian Szegedy,
\newblock ``Explaining and harnessing adversarial examples,''
\newblock {\em arXiv preprint arXiv:1412.6572}, 2014.

\bibitem{madry2017towards}
Aleksander Madry, Aleksandar Makelov, Ludwig Schmidt, Dimitris Tsipras, and
  Adrian Vladu,
\newblock ``Towards deep learning models resistant to adversarial attacks,''
\newblock {\em arXiv preprint arXiv:1706.06083}, 2017.

\bibitem{dong2018boosting}
Yinpeng Dong, Fangzhou Liao, Tianyu Pang, Hang Su, Jun Zhu, Xiaolin Hu, and
  Jianguo Li,
\newblock ``Boosting adversarial attacks with momentum,''
\newblock in {\em Proceedings of the IEEE conference on computer vision and
  pattern recognition}, 2018, pp. 9185--9193.

\bibitem{kreuk2018fooling}
Felix Kreuk, Yossi Adi, Moustapha Cisse, and Joseph Keshet,
\newblock ``Fooling end-to-end speaker verification with adversarial
  examples,''
\newblock in {\em IEEE International Conference on Acoustics, Speech and Signal
  Processing (ICASSP)}. IEEE, 2018, pp. 1962--1966.

\bibitem{moosavi2016deepfool}
Seyed-Mohsen Moosavi-Dezfooli, Alhussein Fawzi, and Pascal Frossard,
\newblock ``Deepfool: a simple and accurate method to fool deep neural
  networks,''
\newblock in {\em Proceedings of the IEEE conference on computer vision and
  pattern recognition}, 2016, pp. 2574--2582.

\bibitem{carlini2017towards}
Nicholas Carlini and David Wagner,
\newblock ``Towards evaluating the robustness of neural networks,''
\newblock in {\em 2017 IEEE Symposium on Security and Privacy (SP)}. IEEE,
  2017, pp. 39--57.

\bibitem{papernot2016distillation}
Nicolas Papernot, Patrick McDaniel, Xi~Wu, Somesh Jha, and Ananthram Swami,
\newblock ``Distillation as a defense to adversarial perturbations against deep
  neural networks,''
\newblock in {\em 2016 IEEE Symposium on Security and Privacy (SP)}. IEEE,
  2016, pp. 582--597.

\bibitem{kannan2018adversarial}
Harini Kannan, Alexey Kurakin, and Ian Goodfellow,
\newblock ``Adversarial logit pairing,''
\newblock {\em arXiv preprint arXiv:1803.06373}, 2018.

\bibitem{taghanaki2019kernelized}
Saeid~Asgari Taghanaki, Kumar Abhishek, Shekoofeh Azizi, and Ghassan Hamarneh,
\newblock ``A kernelized manifold mapping to diminish the effect of adversarial
  perturbations,''
\newblock in {\em Proceedings of the IEEE Conference on Computer Vision and
  Pattern Recognition}, 2019, pp. 11340--11349.

\bibitem{xie2019feature}
Cihang Xie, Yuxin Wu, Laurens van~der Maaten, Alan~L Yuille, and Kaiming He,
\newblock ``Feature denoising for improving adversarial robustness,''
\newblock in {\em Proceedings of the IEEE Conference on Computer Vision and
  Pattern Recognition}, 2019, pp. 501--509.

\bibitem{xu2017feature}
Weilin Xu, David Evans, and Yanjun Qi,
\newblock ``Feature squeezing: Detecting adversarial examples in deep neural
  networks,''
\newblock {\em arXiv preprint arXiv:1704.01155}, 2017.

\bibitem{song2017pixeldefend}
Yang Song, Taesup Kim, Sebastian Nowozin, Stefano Ermon, and Nate Kushman,
\newblock ``Pixeldefend: Leveraging generative models to understand and defend
  against adversarial examples,''
\newblock {\em arXiv preprint arXiv:1710.10766}, 2017.

\bibitem{buckman2018thermometer}
Jacob Buckman, Aurko Roy, Colin Raffel, and Ian Goodfellow,
\newblock ``Thermometer encoding: One hot way to resist adversarial examples,''
\newblock 2018.

\bibitem{jia2019comdefend}
Xiaojun Jia, Xingxing Wei, Xiaochun Cao, and Hassan Foroosh,
\newblock ``Comdefend: An efficient image compression model to defend
  adversarial examples,''
\newblock in {\em Proceedings of the IEEE Conference on Computer Vision and
  Pattern Recognition}, 2019, pp. 6084--6092.

\bibitem{jin2019ape}
Guoqing Jin, Shiwei Shen, Dongming Zhang, Feng Dai, and Yongdong Zhang,
\newblock ``Ape-gan: Adversarial perturbation elimination with gan,''
\newblock in {\em IEEE International Conference on Acoustics, Speech and Signal
  Processing (ICASSP)}. IEEE, 2019, pp. 3842--3846.

\bibitem{ilyas2019adversarial}
Andrew Ilyas, Shibani Santurkar, Dimitris Tsipras, Logan Engstrom, Brandon
  Tran, and Aleksander Madry,
\newblock ``Adversarial examples are not bugs, they are features,''
\newblock {\em arXiv preprint arXiv:1905.02175}, 2019.

\bibitem{sitawarin2019defending}
Chawin Sitawarin and David Wagner,
\newblock ``Defending against adversarial examples with k-nearest neighbor,''
\newblock {\em arXiv preprint arXiv:1906.09525}, 2019.

\bibitem{wong2017provable}
Eric Wong and J~Zico Kolter,
\newblock ``Provable defenses against adversarial examples via the convex outer
  adversarial polytope,''
\newblock {\em arXiv preprint arXiv:1711.00851}, 2017.

\bibitem{oord2018representation}
Aaron van~den Oord, Yazhe Li, and Oriol Vinyals,
\newblock ``Representation learning with contrastive predictive coding,''
\newblock {\em arXiv preprint arXiv:1807.03748}, 2018.

\bibitem{hjelm2018learning}
R~Devon Hjelm, Alex Fedorov, Samuel Lavoie-Marchildon, Karan Grewal, Phil
  Bachman, Adam Trischler, and Yoshua Bengio,
\newblock ``Learning deep representations by mutual information estimation and
  maximization,''
\newblock {\em arXiv preprint arXiv:1808.06670}, 2018.

\bibitem{bachman2019amdim}
Philip Bachman, R~Devon Hjelm, and William Buchwalter,
\newblock ``Learning representations by maximizing mutual information across
  views,''
\newblock {\em arXiv preprint arXiv:1906.00910}, 2019.

\bibitem{hendrycks2019using}
Dan Hendrycks, Mantas Mazeika, Saurav Kadavath, and Dawn Song,
\newblock ``Using self-supervised learning can improve model robustness and
  uncertainty,''
\newblock {\em arXiv preprint arXiv:1906.12340}, 2019.

\bibitem{tschannen2019mutual}
Michael Tschannen, Josip Djolonga, Paul~K Rubenstein, Sylvain Gelly, and Mario
  Lucic,
\newblock ``On mutual information maximization for representation learning,''
\newblock {\em arXiv preprint arXiv:1907.13625}, 2019.

\bibitem{JDH17}
Jeff Johnson, Matthijs Douze, and Herv{\'e} J{\'e}gou,
\newblock ``Billion-scale similarity search with gpus,''
\newblock {\em arXiv preprint arXiv:1702.08734}, 2017.

\bibitem{he2016deep}
Kaiming He, Xiangyu Zhang, Shaoqing Ren, and Jian Sun,
\newblock ``Deep residual learning for image recognition,''
\newblock in {\em Proceedings of the IEEE conference on computer vision and
  pattern recognition}, 2016, pp. 770--778.

\bibitem{wu2018unsupervised}
Zhirong Wu, Yuanjun Xiong, Stella Yu, and Dahua Lin,
\newblock ``Unsupervised feature learning via non-parametric instance-level
  discrimination,''
\newblock {\em arXiv preprint arXiv:1805.01978}, 2018.

\end{thebibliography}

\end{document}